\documentclass[runningheads]{llncs}

\usepackage[mobile]{eccv}

\usepackage{eccvabbrv}

\usepackage{graphicx}
\usepackage{booktabs}

\usepackage[accsupp]{axessibility}  

\usepackage[pagebackref,breaklinks,colorlinks,citecolor=eccvblue]{hyperref}

\usepackage{orcidlink}
\usepackage{graphicx} 
\usepackage{xcolor}
\usepackage{times}
\usepackage{epsfig}
\usepackage{graphicx}
\usepackage{subcaption}
\usepackage{amsmath}
\usepackage{amssymb}
\usepackage{multirow}
\usepackage{colortbl}
\usepackage{booktabs}
\usepackage{url}
\usepackage{eso-pic}
\usepackage{array} 
\usepackage{booktabs}
\usepackage{pifont}
\usepackage{color, colortbl}
\usepackage{hyperref}
\hypersetup{pdfstartview=FitH,
            colorlinks=true,
            linkcolor=red,
            anchorcolor=blue,
            }
\definecolor{mygray}{gray}{.88}
\newcommand{\cmark}{\color{ForestGreen}\ding{51}}%

\begin{document}

\title{Adaptive Calibration: A Unified Conversion Framework of Spiking Neural Networks} 



\author{Ziqing Wang\inst{1}$^*$ \and
Yuetong Fang\inst{1}$^*$  \and
Jiahang Cao\inst{1} \and
Renjing Xu\inst{1}$^\dagger$
}

\authorrunning{Wang,~ et al.}

\institute{The Hong Kong University of Science and Technology (Guangzhou), China
\thanks{$^{*}$ Equal Contribution; $^{\dagger}$ Corresponding Author.}}

\maketitle

\begin{abstract}
Spiking Neural Networks (SNNs) have emerged as a promising energy-efficient alternative to traditional Artificial Neural Networks (ANNs). Despite this, bridging the performance gap with ANNs in practical scenarios remains a significant challenge. This paper focuses on addressing the dual objectives of enhancing the performance and efficiency of SNNs through the established SNN Calibration conversion framework. Inspired by the biological nervous system, we propose a novel Adaptive-Firing Neuron Model (AdaFire) that dynamically adjusts firing patterns across different layers, substantially reducing conversion errors within limited timesteps. Moreover, to meet our efficiency objectives, we propose two novel strategies: an Sensitivity Spike Compression (SSC) technique and an Input-aware Adaptive Timesteps (IAT) technique. These techniques synergistically reduce both energy consumption and latency during the conversion process, thereby enhancing the overall efficiency of SNNs. Extensive experiments demonstrate our approach outperforms state-of-the-art SNNs methods, showcasing superior performance and efficiency in 2D, 3D, and event-driven classification, as well as object detection and segmentation tasks. 
\keywords{Spiking Neural Networks \and Adaptive Calibration \and Neuromorphic Computing}
\end{abstract}

\section{Introduction}

Spiking Neural Networks (SNNs) have gained great attention for their potential to revolutionize the computational efficiency of artificial intelligence systems. Unlike traditional Artificial Neural Networks (ANNs), which compute outputs based on the intensity of neuron activations, SNNs leverage the timing of discrete events or spikes to encode and process information~\cite{maass1997networks}. This temporal dimension allows SNNs to inherently capture the spatio-temporal dynamics of inputs, offering a closer approximation to the way biological neural systems operate~\cite{roy2019towards, vitale2021event, glatz2019adaptive}. When executed on neuromorphic computing devices like Loihi~\cite{daviesLoihiNeuromorphicManycore2018, davies2021advancing} and TrueNorth~\cite{akopyanTruenorthDesignTool2015}, SNNs can achieve substantial improvements in energy efficiency. Recently, researchers have demonstrated SNNs' effectiveness across diverse applications from  classification~\cite{wang2023masked,zhouSpikformerWhenSpiking2022,dengTemporalEfficientTraining2022} to tracking~\cite{zhangSpikingTransformersEventBased2022} and image generation~\cite{cao2024spiking}.

\begin{figure}[t]
	\small
	\hspace{3cm}\includegraphics[width=0.55\linewidth]{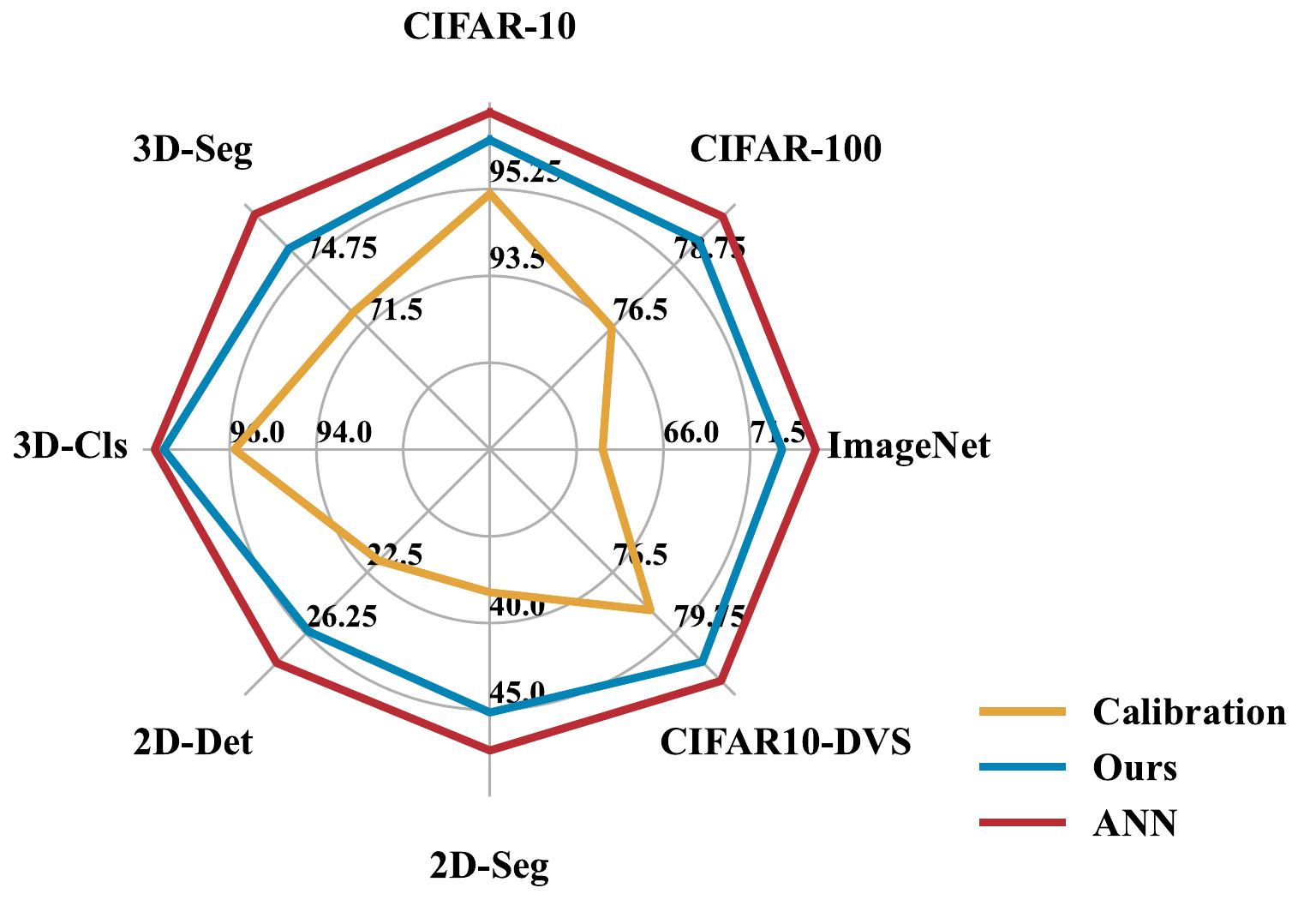}
 	\vskip -0.1in
 \caption{\textbf{Performance comparison on different tasks.} Our method significantly outperforms the traditional Calibration method across all evaluated tasks, effectively narrowing the performance gap with ANNs while requiring limited timesteps.}
	\label{fig:radar}
	\vskip -0.25in
\end{figure}

While SNNs provide notable energy efficiency gains, there still remains a considerable challenge to achieving performance comparable to ANNs in real-world applications. The primary obstacle lies in the inherent non-differentiability of discrete spikes, complicating the training process from scratch~\cite{neftciSurrogateGradientLearning2019}. Current efforts to circumvent this issue primarily revolve around ANN-to-SNN conversion techniques, which involve converting pre-trained ANNs into SNNs. Recent advances in this area focus on reducing conversion errors by substituting the ReLU activation function in ANNs with specially designed activation functions. While these new methods better mimic spiking neuron dynamics during fine-tuning,~\cite{stocklOptimizedSpikingNeurons2021, buOptimalANNSNNConversion2021, dingOptimalAnnsnnConversion2021} however, they require training intermediary surrogate ANNs, extending the overall training period beyond that of traditional ReLU-based ANNs. In addition, these methods require larger timesteps to achieve accuracy levels comparable to ANNs, which increases energy consumption and latency. On the other hand, calibration techniques offer a more straightforward conversion process by aligning the parameters of ReLU-based ANNs with those of SNNs~\cite{liFreeLunchANN2021}, promising a faster conversion process adaptable to various ANNs models. Nevertheless, while calibration does not require re-training, it fails to convert ANNs into high-performance SNNs within an extreme number of timesteps, which undermines the practical deployment of SNNs for low-latency and energy-efficient applications. 

This paper aims to tackle the twin challenges of enhancing the performance and efficiency of SNNs through the established SNNs Calibration conversion framework. Drawing inspiration from the human brain's efficiency, which can execute exaflop-level computations on just 20 watts of power~\cite{versace2010brain, danesh2019synaptic}, we delve into the mechanisms underlying this remarkable efficiency. This extraordinary efficiency stems from the diverse action potential patterns observed in cortical neurons in response to stimuli~\cite{connorsIntrinsicFiringPatterns1990, izhikevichBurstsUnitNeural2003, lismanBurstsUnitNeural1997}. These varied firing patterns, from strong adaptation in regular-spiking cells to the high-frequency, low-adaptation firing of fast-spiking cells, highlight the critical role of adaptive neuronal behaviors in efficient information processing. 

Inspired by these adaptive neuronal behaviors, we present a unified conversion framework aimed at tackling the twin challenges of enhancing the performance and efficiency of SNNs. Unlike conventional approaches that primarily consider a single pattern of neuronal firing, we propose an Adaptive-Firing neuron model (AdaFire) to allocate different firing patterns to different layers, which optimizes the performance. To meet the efficiency objectives, we propose a Sensitivity Spike Compression (SSC) technique as well as an Input-aware Adaptive Timesteps (IAT) technique to reduce both the energy consumption and latency of the conversion process. Collectively, these innovations present a unified conversion framework for enhancing the effectiveness and efficiency of SNNs.

\begin{figure*}[!t]
\centering
\includegraphics[width=0.8\linewidth]{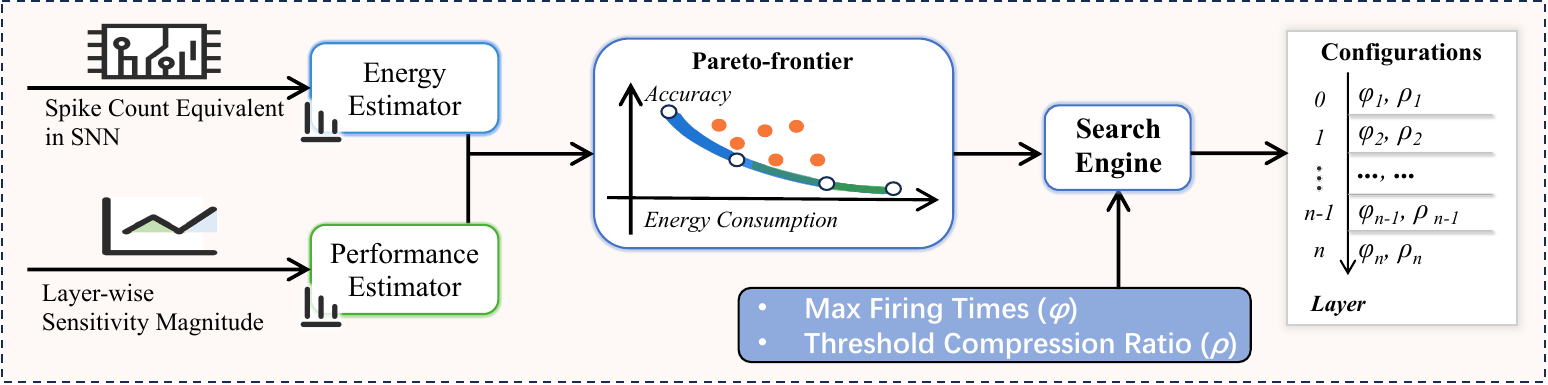}
\caption{The workflow of the Pareto Frontier-driven search algorithm for automatically searching the optimal configurations of each layer in SNNs.} \label{Fig. 2}
\vskip -0.1in
\label{mainfig}
\end{figure*}

In summary, the contributions of our paper are as follows:
\begin{itemize}
    \item We propose an Adaptive-Firing Neuron Model (AdaFire) into the SNN Calibration process to automatically search for the ideal firing patterns of different layers, significantly improving the performance within limited timesteps.
    \item We introduce a Sensitivity Spike Compression (SSC) technique to dynamically adjust the threshold based on the sensitivity of each layer, effectively reducing energy consumption during the conversion process.
    \item We propose an Input-aware Adaptive Timesteps (IAT) technique, which adjusts timesteps dynamically based on the input images, further decreasing both energy consumption and latency.
    \item We have undertaken extensive experiments with our proposed framework across multiple domains, including 2D, 3D, and event-driven classification, object detection, and segmentation tasks. Our experiments reveal that the proposed methodology not only attains state-of-the-art performance but also a remarkable energy savings — up to \textbf{70.1\%}, \textbf{60.3\%}, and \textbf{43.1\%} for the CIFAR-100, CIFAR-10, and ImageNet datasets, respectively. 
\end{itemize}

\section{Related Work and Preliminary}
\label{sec:pre}

\subsection{Spiking Neuron Model}

In SNNs, inputs are transmitted through the neuronal units, typically the Integrate-and-Fire (IF) spiking neuron in ANN-to-SNN conversions~\cite{dingOptimalAnnsnnConversion2021,liFreeLunchANN2021a, buOptimalANNSNNConversion2021a}:
\begin{align}
\label{eq2}
&{u}^{(\ell)}(t+1)={v}^{(\ell)}(t)+{W}^{(\ell)} {s}^{(\ell)}(t) \\
&{v}^{(\ell)}(t+1)={u}^{(\ell)}(t+1)-{s}^{(\ell)}(t+1) \\
&{s}^{(\ell)}(t+1)= \begin{cases}V_{t h}^{(\ell)} & \text { if } {u}^{(\ell)}(t+1) \geq V_{t h}^{(\ell)} \\
0  & otherwise\end{cases}
\end{align}
\noindent where ${u}^{(\ell)}(t+1)$ denotes the membrane potential of neurons before spike generation, ${v}^{(\ell)}(t+1)$ denotes the membrane potential of neurons in layer $\ell$ at time step $t+1$, corresponding to the linear transformation matrix ${W}^{\ell}$, the threshold $\theta^{\ell}$, and binary output spikes ${s}^{\ell}(t+1)$ of current layer $\ell$. In short, a spiking neuron is only active upon receiving or transmitting spikes, thus enabling energy-efficient processing.

\subsection{ANN-to-SNN conversion and SNN Calibration}

The ANN-to-SNN conversion methods~\cite{buOptimalANNSNNConversion2021, dingOptimalAnnsnnConversion2021,liFreeLunchANN2021, dengOptimalConversionConventional2021} involve converting a pre-trained ANNs into SNNs by replacing the \textit{ReLU} activation layers with spiking neurons. Cao et al.~\cite{cao2015spiking} initially exhibited that the \textit{ReLU} neuron is functionally similar to the IF neuron. The average activation value over $T$ timesteps in the IF neuron can be mapped onto that of the \textit{ReLU} neuron directly. However, these methods require the timestep $T$ must be set to infinity, or there could be considerable conversion errors.

To address this issue, Ho et al.~\cite{hoTCLANNtoSNNConversion2021, dingOptimalAnnsnnConversion2021, buOptimalANNSNNConversion2021} proposed to replace the \textit{ReLU} activation function in the original ANNs with a trainable clip function, and find the optimal data-normalization factor through a fine-tuning process to consider both accuracy and latency in the converted SNNs. This clip function is defined as:
\begin{align}
\label{eq3}
\overline{{s}}^{(\ell+1)} & =ClipFloor\left({W}^{(\ell)} \overline{{s}}^{(\ell)}, T, V_{t h}^{(\ell)}\right) \nonumber \\ 
& =\frac{V_{t h}^{(\ell)}}{T} {Clip}\left(\left\lfloor\frac{T}{V_{t h}^{(\ell)}} {W}^{(\ell)} \overline{{s}}^{(\ell)}\right\rfloor, 0, T\right)
\end{align}
\noindent where $\overline{{s}}^{(\ell+1)}$ refers to the averaged spike output over $T$ timesteps in converted SNNs, $\lfloor x \rfloor$ refers to the round down operator. The $Clip$ function limits above but allows below, whereas the $Clip$ function limits a value within a range.

Although the current ANN-to-SNN method is promising, it commonly requires extensive fine-tuning epochs to obtain desired weights and thresholds, consuming a lot of computational resources. Li et al.~\cite{liFreeLunchANN2021} proposed activation transplanting via a layer-wise calibration algorithm aimed at diminishing the discrepancy between the original ANNs and the calibrated SNNs. This spike calibration method determines the optimal threshold by leveraging Eq.~\ref{eq3}:
\begin{equation}
\label{eq4}
\min _{V_{t h}}\left(ClipFloor\left(\overline{{s}}^{(\ell+1)}, T, V_{t h}^{(\ell)}\right)-ReLU\left(\overline{{s}}^{(\ell+1)}\right)\right)^2
\end{equation}
Moreover, to align the outputs of ANNs and SNNs, spike calibration incorporates the expected conversion errors into the bias terms:
\begin{equation}
\label{eq5}
{b}_i^{(\ell)}:={b}_i^{(\ell)}+\mu_i\left({e}^{(\ell+1)}\right)
\end{equation}
where $\mu_i\left({e}^{(\ell+1)}\right)$ computes the spatial mean error between the ANN and SNN outputs in the \( i^{th} \) channel.

\subsection{Spiking Neural Models with Burst-Spike Mechanism}

High-frequency burst-firing neurons, which are commonly found in sensory systems, have been proven to serve distinct functions in information transmission~\cite{krahe2004burst, zeldenrustNeuralCodingBursts2018, connorsIntrinsicFiringPatterns1990, izhikevichBurstsUnitNeural2003, lismanBurstsUnitNeural1997}. Recent research efforts have explored the integration of burst-firing patterns into SNNs~\cite{park2019fast, lan2023efficient, liEfficientAccurateConversion2022}, recognizing their potential to mimic more closely the complex dynamics of biological neural networks. However, these attempts are the uniform application of burst-firing patterns across all layers of SNNs, disregarding the nuanced layer-wise sensitivities inherent to these networks. Moreover, they often overlook a crucial trade-off: while burst firing can improve SNNs performance, it can also lead to increased redundancy and, consequently, higher energy consumption. In contrast, our paper exploits the unique layer-wise sensitivities inherent to SNNs, enabling the adjustment of optimal burst-firing patterns for each layer. We also focus on minimizing energy consumption and latency.

\begin{figure}[tbp]
    \centering
    \begin{minipage}[t]{0.61\textwidth}
        \centering
        \includegraphics[width=\linewidth]{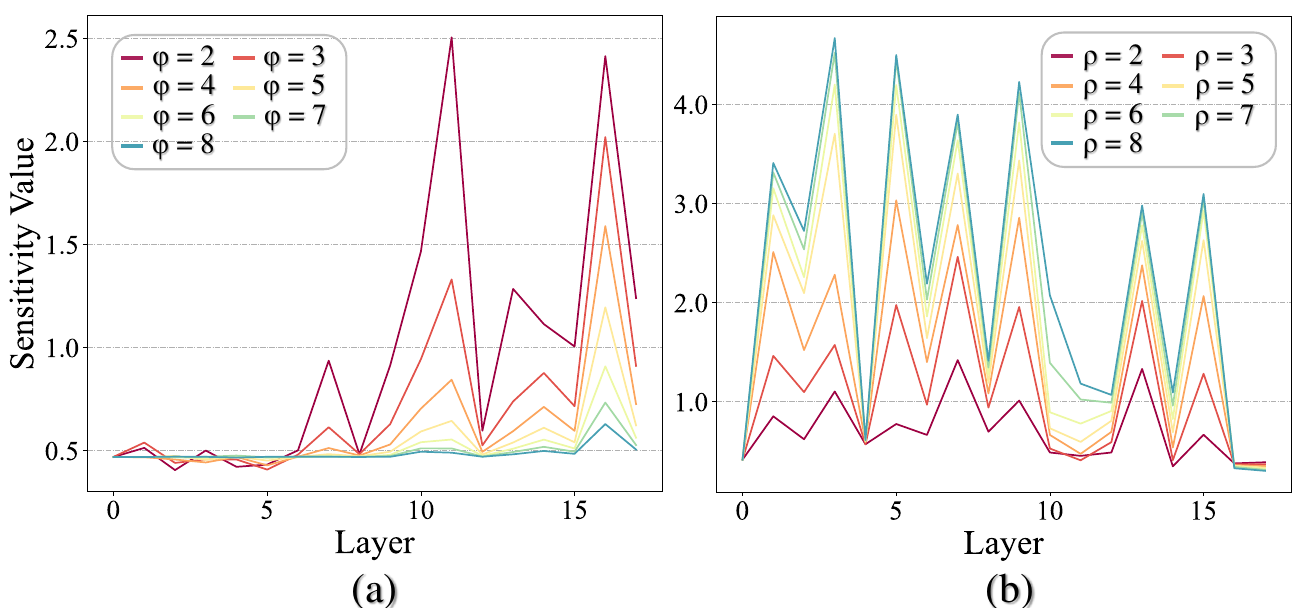}
        \caption{\textbf{Sensitivity result of each layer in ResNet-18.} Subfigure(a): Sensitivity when using different max firing time $\varphi$ for each layer. Subfigure(b): Sensitivity when using different threshold ratios $\rho$ for each layer.}
        \label{fig:sen}
    \end{minipage}\hfill 
    \begin{minipage}[t]{0.37\textwidth}
        \centering
        \raisebox{0.08\height}{\includegraphics[width=\linewidth]{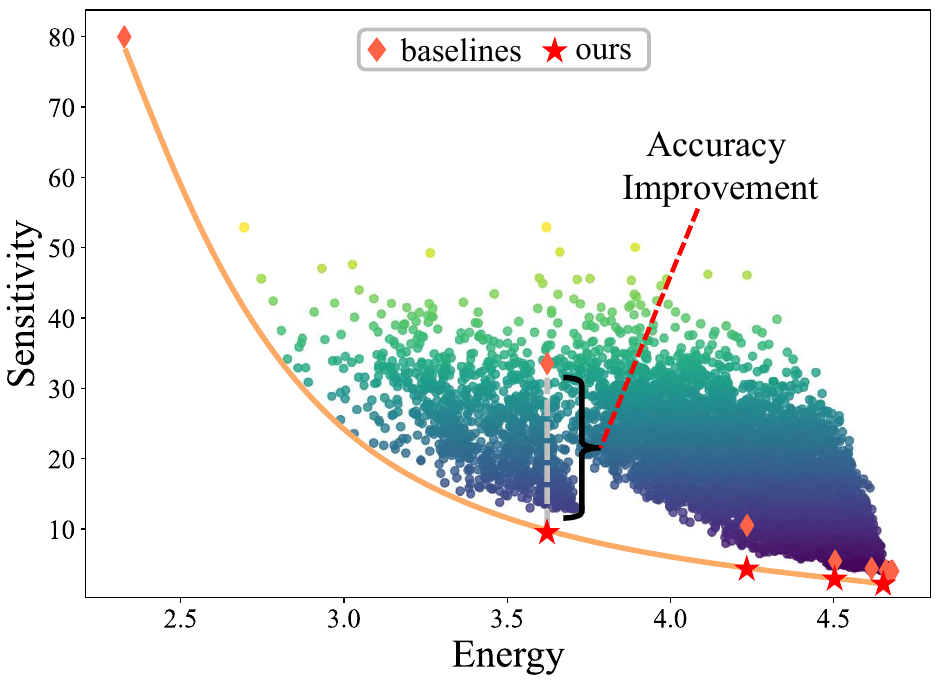}}
        \caption{\textbf{Pareto Frontier Representation.} Each data point represents a distinct layer-specific configuration.}
        \label{fig:pareto}
    \end{minipage}
\end{figure}

\section{Adaptive Calibration}

\subsection{Adaptive-Firing Neuron Model} \label{AdaFire}

In the process of converting Artificial Neural Networks (ANNs) to Spiking Neural Networks (SNNs), a notable challenge arises from the inability of neurons in SNNs to fully mimic the activation outputs of their ANNs counterparts within a limited number of timesteps. This discrepancy primarily stems from the conventional spiking neuron model, which restricts the output range $\overline{{s}}^{(\ell)}$ to \( [0, V_{t h}^{(\ell)}] \), falling short of the maximum activation output observed in ANNs. Such a limitation often results in significant residual information, as depicted in Figure 3, where a gap in activation values between ANNs and SNNs manifests due to residual membrane potential.

\begin{figure}[t]
	\centering
	\small
	\includegraphics[width=0.5\linewidth]{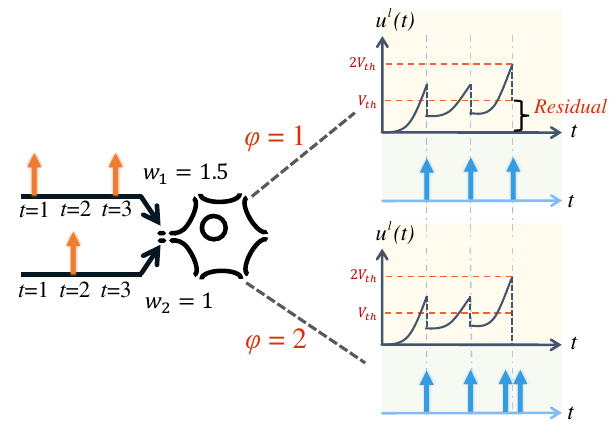}
 \vskip -0.1in
 \caption{\textbf{Adaptive-Firing Mechanism.} Adjusting the maximum firing times \( \varphi \) minimizes residual information, thereby decreasing conversion errors.}
	\label{fig:bs}
\vskip -0.1in
\end{figure}

\vspace{4pt}\noindent\textbf{Burst-Firing Dynamics.} In natural sensory systems, burst-firing neurons, capable of emitting high-frequency spike bursts, have been identified as a mechanism to enhance the fidelity of sensory response transmission~\cite{connorsIntrinsicFiringPatterns1990, izhikevichBurstsUnitNeural2003, lismanBurstsUnitNeural1997}. As shown in Fig.~\ref{fig:bs}, the adoption of a burst-firing model in ANN-to-SNN conversion holds the promise of facilitating more efficient information transmission and reducing the residual information, thereby potentially lowering the conversion error~\cite{park2019fast, lan2023efficient, liEfficientAccurateConversion2022}. To this end, we reconsider the dynamics of a spiking neuron to improve the spike calibration method. Different from prior studies, the burst-spike burst-firing neuron model allows up to \( \varphi \) spikes per timestep rather than merely one spike. This modification effectively expands the potential range of neuronal activation output $\overline{{s}}^{(\ell)}$ to \( [0, V_{t h}^{(\ell-1) } \times \varphi] \). Consequently, the relationship of the activation output between ANNs and the converted SNNs in Eq.~\ref{eq4} then becomes:
\begin{align}
\label{eq6}
\overline{{s}}^{(\ell+1)} & = ClipFloor\left({W}^{(\ell)} \overline{{s}}^{(\ell)}, T, V_{t h}^{(\ell)}, \varphi^{(\ell)}\right) \nonumber \\ 
& = \frac{V_{t h}^{(\ell)}}{T} Clip\left(\left\lfloor\frac{T}{V_{t h}^{(\ell)}} {W}^{(\ell)} \overline{{s}}^{(\ell)}\right\rfloor, 0, T \times \varphi\right)
\end{align}
Correspondingly, we modify the Eq.~\ref{eq5} used to determine the optimal threshold as: 
\begin{equation}
\label{eq7}
\min _{V_{t h}}\left(ClipFloor\left(\overline{{s}}^{(\ell+1)}, T \times \varphi, V_{t h}^{(\ell)}\right)-ReLU\left(\overline{{s}}^{(\ell+1)}\right)\right)^2
\end{equation}

\vspace{4pt}\noindent\textbf{Layer-Specific Firing Patterns Adaptation.}
Uniformly applying burst-firing capabilities, denoted as \( \varphi \), across all layers in SNNs may not yield optimal results. This premise is inspired by the natural variability in firing patterns of biological neurons, which are tailored to their specific functional roles~\cite{mochizukiSimilarityNeuronalFiring2016, zeldenrustNeuralCodingBursts2018}. Our observation into the adaptability of SNNs layers to changes in \( \varphi \) reveals a significant insight, as depicted in Fig.~\ref{fig:sen}:

\vspace{4pt}\noindent\textbf{Observation 1:} \textit{The sensitivity of each layer to changes in \(\varphi\) varies significantly.}

\vspace{4pt}\noindent Leveraging this observation, we argue that layers with higher sensitivity to 
\(\varphi\) variations should be allocated a greater range of firing patterns. This insight leads us to the development of the Adaptive-Firing Neuron Model (AdaFire), which judiciously considers both the sensitivity of each layer to firing pattern changes and the overarching goal of minimizing energy consumption.

\vspace{4pt}\noindent\textbf{Performance Metric.}
We use sensitivity to estimate the performance of SNNs. We demonstrate that the smaller the value of sensitivity, the higher the SNNs performance (shown in Appendix). To quantify layer sensitivity, we employ Kullback-Leibler (KL) divergence~\cite{caiZeroqNovelZero2020}, a measure that quantifies the difference in output distributions between the ANN and SNN configurations for each layer. Specifically, the sensitivity metric for the \(i\)-th layer with respect to parameter \(k\) is given by:
\begin{equation}
\label{eq8}
S_i(k) = \frac{1}{N} \sum_{j=1}^{N} \mathrm{KL}\left(\mathcal{M}\left(ANN_i; x_j\right), \mathcal{M}\left(SNN_i(k) ; x_j\right)\right)
\end{equation}
Here, a lower \(S_i(k)\) value indicates that the SNNs model's output closely aligns with that of the ANNs model for the chosen \(k\) at layer \(i\), signifying a lesser sensitivity to changes in \( \varphi \), and vice versa. Notably, our analysis (refer to Fig.~\ref{fig:sen}(a)) demonstrates that early layers tend to be less sensitive to \( \varphi \) modifications, while deeper layers require an increased \( \varphi \) to accommodate their increased sensitivity.

\vspace{4pt}\noindent\textbf{Efficiency Metric.}
For assessing SNNs efficiency and energy consumption, we draw upon established methodologies~\cite{wang2023masked, dingOptimalAnnsnnConversion2021, cao2015spiking}, calculating energy based on the total number of spikes and their associated energy cost per spike, \( \alpha \) Joules. Energy consumption over a 1 ms timestep is thus represented as:
\begin{equation}
   E = \frac{\text{total spikes}}{1 \times 10^{-3}} \times \alpha \quad (\text{in Watts})
\label{eq9}
\end{equation}
This formulation acknowledges the direct correlation between spike counts and energy expenditure in SNNs, thereby serving as a practical metric of energy consumption.

\vspace{4pt}\noindent\textbf{Pareto Frontier Driven Search Algorithm.}
Optimizing the layer-specific maximum number of firing patterns \( \varphi \) is a non-trivial challenge. For an SNN model with \( L \) layers and \( n \) configurations per layer, the possible combinations total \( n^L \). This count grows exponentially with increasing layers. To address the vast search space, we conduct layer-wise searches based on the dynamic programming algorithm. Our purpose is to find the optimal combination that minimizes the overall sensitivity value \( S_{\text{sum}} \) under a predefined energy budget \( E_{\text{target}} \). We achieve this by leveraging the Pareto frontier approach, which guides us in pinpointing configurations that offer the best trade-off between sensitivity reduction and energy consumption. Formally, the optimization problem is defined as:
\begin{equation}
\label{eq10}
\min _{\left\{k_i\right\}_{i=1}^L} S_{\text{sum}} = \sum_{i=1}^L S_i\left(k_i\right), \quad\sum_{i=1}^L E_i \leq E_{\text{target}}
\end{equation}
where \( k_i \) symbolizes the chosen configuration for the \( i^{th} \) layer, and \( E_i \) refers to the estimated energy consumption the same layer. We operate under the assumption that the sensitivity of each layer to its configuration is independent of the configurations of other layers. This assumption enables a simplification of the model's performance optimization to a summation of individual layer sensitivities. By leveraging the dynamic programming algorithm, the optimal combination of \( \varphi \) for various \( E_{\text{target}} \) values can be concurrently determined. As shown in Fig.~\ref{fig:pareto}, our method effectively balances the trade-off between energy consumption and sensitivity, achieving superior performance over baseline approaches that don't employ a systematic search strategy.

\begin{figure}[t]
	\centering
	\small
	\includegraphics[width=0.5\linewidth]{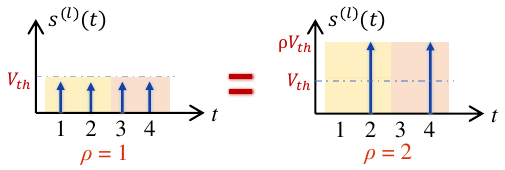}
 \vskip -0.1in
 \caption{\textbf{Spike Compression Mechanism.} Our approach enables the compression of regular spikes, preserving information integrity.}
	\label{fig:thres}
\vskip -0.1in
\end{figure}

\subsection{Sensitivity Spike Compression}
In this section, we focus on improving the efficiency during the conversion process. We devise a method to inhibit spike generation, thereby reducing the number of spikes and energy consumption.

\vspace{4pt}\noindent\textbf{Adaptive Threshold.}
As depicted in Fig.~\ref{fig:thres}, when a neuron emits spikes at regular intervals, consecutive spikes can be compressed by a singular, double-amplitude spike without losing the original timing information. This process can be mathematically represented as:
\begin{equation}
V_{t h}^{(\ell)}=\rho^{(\ell)} \cdot v_{t h}^{(\ell)}
\label{eq11}
\end{equation}
where \(\rho^{(\ell)} \) refers to the threshold amplification ratio and \( v_{t h}^{(\ell)} \) signifies the initial threshold of layer \( \ell \). The subsequent spike output of an IF neuron can be described by:
\begin{align}
\label{eq12}
{s}^{(\ell+1)}(t) &= \begin{cases} 
\rho^{(\ell)} \cdot V_{t h}^{(\ell)} & \text{if } {u}^{(\ell)}(t+1) \geq \rho^{(\ell)} \cdot V_{t h}^{(\ell)} \\
0 & otherwise
\end{cases}
\end{align}
Subsequently, the updated firing rate for SNN output is:
\begin{equation}
{r}^{(\ell+1)} = \sum_{i=1}^n {W}_i^{(\ell)} \frac{\sum_{t=1}^T {s}_i^{(\ell)}(t) \cdot \rho^{(\ell)}}{T}
\label{eq13}
\end{equation}
This method effectively decreases the spike generation while ensuring that the quantity of information conveyed through each neuron is amplified by the factor \(\rho^{(\ell)}\), maintaining the integrity of information transmission across layers.

\vspace{4pt}\noindent\textbf{Adaptive Threshold Search Algorithm.}
Applying threshold compression naively could, however, lead to significant performance degradation, particularly for irregular spike trains where spike compression could result in data loss. To mitigate this, we propose the Sensitivity Spike Compression (SSC) method. SSC evaluates the impact of threshold ratio (\(\rho\)) modifications on output variability. For each layer, the objective is to identify an optimal \(\rho\) that reduces spike generation with minimal impact on accuracy. An important insight emerged from our analysis, as shown in Fig.~\ref{fig:sen}:

\vspace{4pt}\noindent\textbf{Observation 2:} \textit{The sensitivity of each layer to changes in \(\rho\) varies significantly.}

\vspace{4pt}\noindent Intuitively, applying higher \(\rho\) values to layers with lower sensitivity can lead to a pronounced reduction in the network's overall spike. To identify the most effective \(\rho\) configuration, we adapt the search algorithm outlined in Section~\ref{AdaFire}, modifying its objective to pinpoint the lowest possible energy consumption for a predefined sensitivity budget \(S_{\text{target}}\), which denotes the permissible limit for performance decline:
\begin{equation}
\min_{\{\rho_i\}_{i=1}^L} E_{\text{sum}} = \sum_{i=1}^L E_i(\rho_i), \quad \sum_{i=1}^L S_i \leq S_{\text{target}}.
\label{eq14}
\end{equation}
Empirical results confirm that the SSC approach significantly lowers energy consumption while only minimally affecting performance.

\subsection{Input-aware Adaptive Timesteps}
In traditional SNNs configurations, the number of timesteps is often set as a fixed hyper-parameter. However, this fixed approach overlooks the potential benefits of dynamically adjusting timesteps to suit the unique demands of each input image. Recent observations suggest the potential of SNNs to adjust timesteps dynamically, based on the unique characteristics of individual input images~\cite{li2024seenn}.

\vspace{4pt}\noindent\textbf{Entropy as a Confidence Measure.}
Inspired by~\cite{teerapittayanon2016branchynet, guo2017calibration}, we employ entropy as a confidence measure for predictions at each timestep. Formally, the entropy \(H(p)\) of a probability distribution \(p\) over a label space \(\mathcal{Y}\) is defined as:
\begin{equation}
    H(p) := \sum_{y \in \mathcal{Y}} p_y \log p_y,
    \label{eq15}
\end{equation}
where \(p_y\) represents the probability of label \(y\) within the distribution \(p\). 

\vspace{4pt}\noindent\textbf{Dynamic Timestep Adjustment Mechanism.}
We adopt a threshold-based mechanism, with a predefined threshold \(\alpha\), to evaluate the confidence score dynamically. The SNN exits the inference process when the score surpasses \(\alpha\), thus optimizing the balance between accuracy and inference time:
\begin{equation}
\begin{aligned}
\mathbf{P}_{\text{SNN}}: \min _{\alpha \in S} & \mathbb{E}_{(\mathbf{x}, y) \sim \mathcal{D}}[-a(\mathbf{x}, y, \alpha)], \\
\text { s.t. } & \mathbb{E}_{(\mathbf{x}, y) \sim \mathcal{D}}[b(\mathbf{x}, \alpha)] \leq \Gamma,
\end{aligned}
    \label{eq16}
\end{equation}
where \(\Gamma\) is the predefined average latency requirement, and \(a\) and \(b\) represent accuracy and latency functions, respectively. Given the practical challenge of solving this optimization problem due to its non-convex nature, we seek to find a good approximate solution through empirical approximation.

\vspace{4pt}\noindent\textbf{Adaptive Threshold.}
Unlike prior works that utilize a single fixed threshold \(\alpha\) for all timesteps~\cite{li2024seenn}, we recognize the varied contribution of each timestep to the network's final accuracy. We meticulously analyze the entropy distribution across timesteps for the training dataset, revealing significant differences at each timestep. This discovery highlights the inadequacy of applying a uniform threshold and sets the stage for our proposed input-aware adaptive timestep technique. The threshold of confidence score at each timestep is determined by the following formula:
\begin{equation}
\label{eq17}
\alpha_{t}=\alpha_{base}+\beta e^{-\frac{\bar{E}_t-\bar{E}_{min}}{\delta}}
\end{equation}
where $\alpha_{base}$ is the base threshold, $\beta$ is the scaling factor, $\delta$ represents the decay constant, $\bar{E}_t$ denotes the average entropy of the network's output distribution at timestep $t$, and $\bar{E}_{min}$ the minimum average entropy observed across all timesteps. This formulation allows for a dynamic threshold adjustment: a higher average entropy at a given timestep, indicating high confidence in the output, warrants a lower threshold to expedite inference. Conversely, a low entropy value, especially at initial timesteps, suggests the necessity for a higher threshold to prevent premature exits that could detrimentally affect accuracy. The detailed methodology, including the pseudo-code for this algorithm, is provided in the Appendix.

\section{Experiment}
We systematically evaluate our unified conversion framework across a diverse benchmarks, including tasks in 2D and 3D classification, event-driven classification, object detection, and segmentation. This extensive experiment aims to demonstrate the framework's superior performance and efficiency. In subsequent sections, we set $\varphi$ to 4 by default. Comprehensive details on our experimental setup are provided in the Appendix.
\vskip -0.1in
\begin{table}[!t]
\setlength\tabcolsep{6pt} 
\renewcommand{\arraystretch}{1}

\centering
\scriptsize
\caption{Performance comparison between the proposed model and the state-of-the-art models on the ImageNet dataset.}
\vskip -0.1in
\label{tb2}
\begin{tabular}{ccccccc}
\toprule
\textbf{Architecture}               & \textbf{Method} & \textbf{ANN} & \textbf{T=8}     & \textbf{T=16}    & \textbf{T=32}    & \textbf{T=64}    \\ \midrule
\multirow{5}{*}{\textbf{VGG-16}}    & OPT~\cite{dengOptimalConversionConventional2021}\textsuperscript{ICLR} & 75.36            & - & - & 0.11            & 0.12            \\
                                    & SNM~\cite{wangSignedNeuronMemory2022}\textsuperscript{IJCAI}   & 73.18     & - & - & 64.78            & 71.50          \\
                                    & QCFS~\cite{buOptimalANNSNNConversion2021}\textsuperscript{ICLR} & 74.39           & - & - & 68.47            & 72.85            \\
                                    & Calibration~\cite{liFreeLunchANN2021}\textsuperscript{ICML} & 75.36    & 25.33 & 43.99 & 62.14 & 65.56 \\ 
                                    & \cellcolor{mygray}\textbf{AdaFire (Ours)} & \cellcolor{mygray} 75.36 & \cellcolor{mygray}\textbf{73.53} & \cellcolor{mygray}\textbf{74.25}  & \cellcolor{mygray}\textbf{74.98} & \cellcolor{mygray}\textbf{75.22} \\ \midrule
\multirow{4}{*}{\textbf{ResNet-34}} & OPT~\cite{dengOptimalConversionConventional2021}\textsuperscript{ICLR} & 75.66           & - & - & 0.11            & 0.12            \\
                                    & QCFS~\cite{buOptimalANNSNNConversion2021}\textsuperscript{ICLR}  & 74.32           & - & - & 69.37            & 72.35            \\
                                    & Calibration~\cite{liFreeLunchANN2021}\textsuperscript{ICML} & 75.66     & 0.25 & 34.91 & 61.43 & 69.53 \\
                                    & \cellcolor{mygray}\textbf{AdaFire (Ours)} & \cellcolor{mygray} 75.66  & \cellcolor{mygray}   \textbf{72.96} & \cellcolor{mygray}\textbf{73.85}  &\cellcolor{mygray} \textbf{75.04}  & \cellcolor{mygray}\textbf{75.38} \\
                    \bottomrule 
\end{tabular}
\vskip -0.15in
\end{table}

\subsection{Effectiveness of Adaptive-Firing Neuron Model}
\noindent\textbf{Performance on Static Classification.}
We utilize the ImageNet~\cite{dengImagenetLargescaleHierarchical2009} dataset to evaluate effectiveness of our method. Tab.~\ref{tb2} offers an exhaustive comparison between our method and the current state-of-the-art SNNs conversion techniques. This comparison highlights the unique capability of our method to maintain high accuracy levels even under limited timesteps. 
Typically, methods like OPT~\cite{dengOptimalConversionConventional2021} and QCFS~\cite{buOptimalANNSNNConversion2021} exhibit accuracy reductions at fewer timesteps, as they depend on elongated timesteps to manage activation mismatch. Contrarily, our method, empowered by the Adaptive-Firing Neuron Model (AdaFire), minimizes information loss during the conversion phase, thereby bolstering accuracy. To ensure a fair evaluation, we align our model operating at $T=8$ with competitors set at $T=32$, ensuring they are compared at equivalent energy consumption levels. The results in VGG16 show that our method surpasses Calibration which is the base framework of our method for about \textbf{11.39\%}. Additionally, our method outperforms QCFS~\cite{buOptimalANNSNNConversion2021} and SNM~\cite{wangSignedNeuronMemory2022} by margins of \textbf{5.06\%} and \textbf{8.75\%} respectively. 

%

\begin{table}[!t]
\setlength\tabcolsep{6pt} 
\renewcommand{\arraystretch}{1}

\centering
\scriptsize
\caption{Performance comparison between the proposed model and the state-of-the-art models on different neuromorphic datasets.}
\vskip -0.1in
\label{tb1}
\begin{tabular}{cccc}
\toprule 
\textbf{Dataset}              & \textbf{Model}           & \textbf{Timesteps}          &                                  \textbf{Accuracy (\%)} \\ \midrule
\multirow{6}{*}{\textbf{CIFAR10-DVS}}  
                              & TA-SNN~\cite{yaoTemporalwiseAttentionSpiking2021}\textsuperscript{ICCV}        & 10                             & 72.00                \\
                              & PLIF~\cite{fangIncorporatingLearnableMembrane2021}\textsuperscript{ICCV}           & 20                               & 74.80             \\
                              & Dspkie~\cite{liDifferentiableSpikeRethinking2021}\textsuperscript{NeurIPS}        & 10                              & 75.40              \\
                              & DSR ~\cite{mengTrainingHighPerformanceLowLatency2022}\textsuperscript{CVPR}           & 10                        & 77.30              \\
                              & Spikformer ~\cite{zhouSpikformerWhenSpiking2022}\textsuperscript{ICLR}   & 10                                & 80.90              \\
                              & \cellcolor{mygray}\textbf{AdaFire (Ours)}    & \cellcolor{mygray}\textbf{8}                                               & \cellcolor{mygray}\textbf{81.25}    \\ 
                              \midrule
\multirow{3}{*}{\textbf{N-Caltech101}} 
                              & SALT~\cite{kimOptimizingDeeperSpiking2021a}\textsuperscript{NN}          & 20                                & 55.00                \\
                              & NDA~\cite{liNeuromorphicDataAugmentation2022}\textsuperscript{ECCV}            & 10                                 & 83.70              \\
                              & \cellcolor{mygray}\textbf{AdaFire (Ours)}     &\cellcolor{mygray} \textbf{8}                          & \cellcolor{mygray}\textbf{85.21}    \\ 
                              \midrule
\multirow{3}{*}{\textbf{N-Cars}}       
                              & CarSNN~\cite{vialeCarsnnEfficientSpiking2021}\textsuperscript{IJCNN}        & 10                           & 86.00                \\
                              & NDA ~\cite{liNeuromorphicDataAugmentation2022}\textsuperscript{ECCV}          & 10                               & 91.90              \\
                              & \cellcolor{mygray}\textbf{AdaFire (Ours)}      & \cellcolor{mygray}\textbf{8}                            & \cellcolor{mygray}\textbf{96.24}   \\ 
                              \midrule
\multirow{3}{*}{\textbf{Action Recognition}}       
                              & STCA~\cite{guSTCASpatiotemporalCredit2019}\textsuperscript{IJCAI}        & 10                           & 71.20                \\
                              & Mb-SNN ~\cite{liuEventbasedActionRecognition2021a}\textsuperscript{IJCAI}          & 10                               & 78.10             \\
                              & \cellcolor{mygray}\textbf{AdaFire (Ours)}      & \cellcolor{mygray}\textbf{8}                            &\cellcolor{mygray} \textbf{88.21}   \\ 
                                              \bottomrule  
\end{tabular}
\vskip -0.1in
\end{table}

\vspace{4pt}\noindent\textbf{Performance on Event-driven Classification.}
As shown in Tab.~\ref{tb1}, our approach demonstrated superior performance over other leading SNN techniques across various neuromorphic datasets within limited timesteps. Datasets like CIFAR10-DVS, N-Caltech101, and N-Cars were derived from static datasets through event-based cameras. Our results show that our approach, enhanced with the AdaFire technique, consistently outperforms other leading SNN models. For instance, our method significantly outperforms the PLIF model~\cite{fangIncorporatingLearnableMembrane2021}, which uses 20 timesteps, by \textbf{6.45\%} using only 8 timesteps. Moreover, in the domain of Action Recognition which encapsulates sequential human actions recorded with event-based cameras, our model achieves an impressive top-1 accuracy of \textbf{88.21\%}. These results, markedly better than alternatives, underscore our method's adaptability to diverse neuromorphic datasets.

\begin{table}[t]
\renewcommand{\arraystretch}{1}
\setlength\tabcolsep{6pt}
\centering
\scriptsize
\caption{Performance comparison for object detection on PASCAL VOC 2012 and MS COCO 2017 datasets. mAP represents the mean Average Precision.}
\vskip -0.1in
\label{tab:detection}
\begin{tabular}{cccccc}
\toprule
\textbf{Dataset}               & \textbf{Method}         & \textbf{Architecture} & \textbf{ANN} & \textbf{Timesteps} & \textbf{mAP}   \\ \midrule
\multirow{4}{*}{\textbf{VOC}}  & Spiking-YOLO~\cite{kimSpikingyoloSpikingNeural2020}\textsuperscript{AAAI}            & Tiny YOLO      & 53.01        & 8000       & 51.83          \\
                               & B-Spiking-YOLO~\cite{kimFastAccurateObject2020}\textsuperscript{Access}          & Tiny YOLO      & 53.01        & 5000       & 51.44          \\
                               & Calibration~\cite{liFreeLunchANN2021}\textsuperscript{ICML}             & YOLOv2         & 76.16        & 128        & 67.65          \\
                               & \cellcolor{mygray}\textbf{AdaFire (Ours)} & \cellcolor{mygray}YOLOv2         & \cellcolor{mygray}76.16        & \cellcolor{mygray}16         & \cellcolor{mygray}\textbf{75.17} \\ \midrule
\multirow{4}{*}{\textbf{COCO}} & Spiking-YOLO~\cite{kimSpikingyoloSpikingNeural2020}\textsuperscript{AAAI}             & Tiny YOLO      & 26.24        & 8000       & 25.66          \\
                               & B-Spiking-YOLO~\cite{kimFastAccurateObject2020}\textsuperscript{Access}           & Tiny YOLO      & 26.24        & 5000       & 25.78          \\
                               & Calibration~\cite{liFreeLunchANN2021}\textsuperscript{ICML}             & YOLOv2         & 29.46        & 128        & 21.79          \\
                               & \cellcolor{mygray}\textbf{AdaFire (Ours)} & \cellcolor{mygray}YOLOv2         &\cellcolor{mygray} 29.46        & \cellcolor{mygray}16         & \cellcolor{mygray}\textbf{28.04} \\
\bottomrule
\end{tabular}
\vskip -0.25in
\end{table}

\vspace{4pt}\noindent\textbf{Performance on Object Detection.}
Our study delves deeper into object detection advancements, utilizing the widely recognized PASCAL VOC 2012~\cite{everingham2010pascal} and MS COCO 2017~\cite{linMicrosoftCocoCommon2014} datasets for evaluation. In our analysis, we benchmark the performance of our proposed method against well-established models. As show in Tab.~\ref{tab:detection}, our experiments on the COCO dataset reveals a marked improvement in model efficiency. Notably, Spiking-YOLO~\cite{kimSpikingyoloSpikingNeural2020} achieves a mean Average Precision (mAP) of 26.23\% over an extensive computational budget of 8000 timesteps. In contrast, our method significantly outperforms this with an mAP of \textbf{28.04\%} while requiring merely 16 timesteps. This dramatic reduction in timesteps translates to a speed-up of approximately \textbf{500$\times$} compared to Spiking-YOLO~\cite{kimSpikingyoloSpikingNeural2020}. Such an enhancement not only underscores our method's superior accuracy but also its feasibility for real-time application scenarios.

\begin{table}[t]
\centering
\begin{minipage}[t]{0.55\linewidth} 
\renewcommand{\arraystretch}{1}
\setlength\tabcolsep{4pt} 
\scriptsize
\centering
\caption{Performance comparison for Semantic Segmentation.}
\vskip -0.1in
\label{tab:2dseg}
\begin{tabular}{cccccc}
\toprule
\textbf{Dataset}               & \textbf{Method}         & \textbf{Arch.} & \textbf{ANN} & \textbf{T}  & \textbf{mAP}   \\ \midrule
\multirow{2}{*}{\textbf{VOC}}  & Calibration~\cite{liFreeLunchANN2021}             & ResNet50       & 73.36        & 128         & 69.11          \\
                               & \cellcolor{mygray}\textbf{AdaFire (Ours)} & \cellcolor{mygray}ResNet50       & \cellcolor{mygray}73.36        & \cellcolor{mygray}\textbf{16} & \cellcolor{mygray}\textbf{72.17} \\
                               \midrule
\multirow{2}{*}{\textbf{COCO}} & Calibration~\cite{liFreeLunchANN2021}             & ResNet50       & 47.34        & 128         & 38.23          \\
                               & \cellcolor{mygray}\textbf{AdaFire (Ours)} & \cellcolor{mygray}ResNet50       &\cellcolor{mygray} 47.34        & \cellcolor{mygray}\textbf{16} & \cellcolor{mygray}\textbf{45.15} \\
\bottomrule
\end{tabular}
\end{minipage}
\begin{minipage}[t]{0.44\linewidth}
\renewcommand{\arraystretch}{1}
\setlength\tabcolsep{4pt} 
\centering
\scriptsize
\caption{Performance comparison for 3D Classification on the ShapeNet dataset.}
\vskip -0.1in
\label{tab:3dcls}
\begin{tabular}{cccc}
\toprule
\textbf{Method} & \textbf{Arch.} & \textbf{T} & \textbf{Acc.} \\
\midrule
ANN & PointNet & / & 97.73 \\
Calibration~\cite{liFreeLunchANN2021} & PointNet & 64 & 95.89 \\
\rowcolor{mygray}\textbf{AdaFire (Ours)} & PointNet & \textbf{16} & \textbf{97.52} \\
\bottomrule
\end{tabular}
\end{minipage}
\vskip -0.1in
\end{table}

\vspace{4pt}\noindent\textbf{Performance on Semantic Segmentation.}
We extend the exploration to semantic segmentation task, utilizing the benchmark PASCAL VOC 2012 and MS COCO 2017 datasets. Semantic segmentation has seen limited exploration in SNNs, presenting a unique opportunity. As show in Tab.~\ref{tab:2dseg}, our AdaFire model, designed as an advancement over the conventional Calibration baseline, demonstrates our capability to significantly enhance mAP across both datasets while substantially reducing the timesteps. This achievement not only confirms the effectiveness of our model but also marks a pioneering step in applying SNNs to semantic segmentation.

\vspace{4pt}\noindent\textbf{Performance on 3D Classification.}
The exploration of SNNs in 3D task domains remains relatively nascent, despite the growing ubiquity of 3D technologies across a wide array of applications such as remote sensing, augmented/virtual reality (AR/VR), robotics, and autonomous driving. In response to this need, our study extends the application to the task of 3D point cloud classification. We conduct our evaluation using the ShapeNet dataset~\cite{yi2016scalable}, employing PointNet~\cite{qi2017pointnet} as the architectural backbone. The comparative analysis in Tab.~\ref{tab:3dcls} reveals that our method with merely 16 timesteps, achieves a notable improvement in classification accuracy, outperforming the Calibration baseline by 1.63\%.

\begin{table}[t]
\renewcommand{\arraystretch}{1}
\setlength\tabcolsep{6pt}
\centering
\scriptsize
\caption{Performance comparison for 3D Part Segmentation on the ShapeNet dataset. Our method uses $T$ = 16 and baseline uses $T$= 64.}
\vskip -0.1in
\label{tab:3dseg}
\begin{tabular}{c|c|cccccccc}
\hline
\textbf{Method} & \textbf{Mean} & \textbf{Aero} & \textbf{Bag} & \textbf{Cap} & \textbf{Car} & \textbf{Chair}  & \textbf{Guitar} & \textbf{Knife} & \textbf{Earphone} \\
\hline
ANN & 77.46 & 81.55 & 78.74 & 71.87 & 75.15 & 89.1  & 89.22 & 83.81 & 69.55 \\
Calibration~\cite{liFreeLunchANN2021} & 72.25 & 74.11 & 78.07 & 70.15 & 62.53 & 79.48  & 83.52 & 77.88 & 66.97 \\
\rowcolor{mygray}\textbf{AdaFire (Ours)} & \textbf{75.65} & \textbf{78.99} & \textbf{78.91} & \textbf{70.89} & \textbf{71.95} & \textbf{88.17}  & \textbf{86.03} & \textbf{82.85}  & \textbf{67.27}\\
\hline
\textbf{Method} & \textbf{Mean} & \textbf{Lamp} & \textbf{Laptop} & \textbf{Motor} & \textbf{Mug} & \textbf{Pistol} & \textbf{Rocket} & \textbf{Table} & \textbf{Skateboard}  \\
\hline
ANN & 77.46 & 80.74 & 94.54 & 60.95 & 85.58 & 80.68 & 44.94  & 81.45 & 71.47 \\
Calibration~\cite{liFreeLunchANN2021} & 72.25 & 76.53 & 92.45 & 56.27 & 78.91 & 76.58 & 42.54  & 76.91  & 63.1\\
\rowcolor{mygray}\textbf{AdaFire (Ours)} & \textbf{75.65} & \textbf{79.13} & \textbf{92.47} & \textbf{61.64} & \textbf{85.23} & \textbf{80.11} & \textbf{43.16}  & \textbf{78.32} & \textbf{65.3} \\
\hline
\end{tabular}
\vskip -0.2in
\end{table}

\vspace{4pt}\noindent\textbf{Performance on 3D Part Segmentation.}
We extend the application of SNNs to the domain of 3D Part Segmentation, marking a pioneering effort in this area. Part segmentation represents a nuanced challenge within 3D recognition tasks, requiring the assignment of specific part category labels (e.g., chair leg, cup handle) to each point or facet of a given 3D scan or mesh model. The results, as detailed in Tab.~\ref{tab:3dseg}, illustrate a significant advancement over the established Calibration baseline. Our method achieves a performance improvement of \textbf{3.4\%} with only 16 timesteps. This result underscores the potential of SNNs in handling complex 3D tasks with higher energy efficiency and lower latency.

\begin{figure}[t]
	\centering
	\small
	\includegraphics[width=1\linewidth]{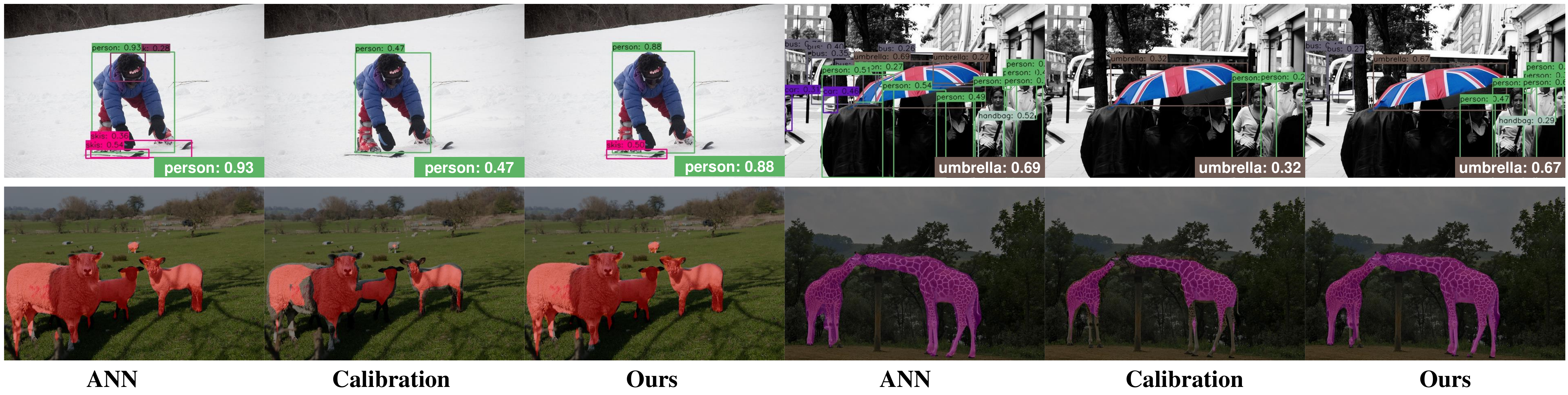}
 \vskip -0.1in
\caption{\textbf{Visualization on the COCO dataset.} The first row displays object detection results, while the second row showcases semantic segmentation results.}
	\label{fig:coco}
\vskip -0.1in
\end{figure}

\vspace{4pt}\noindent\textbf{Visualization.}
Fig~.\ref{fig:coco} visualizes the results of our object detection and semantic segmentation tasks, showcasing the significant enhancements using our method. In the domain of object detection, our AdaFire model substantially increases the accuracy and reliability of recognition. For example, where the Calibration method fails to detect a sled within 128 timesteps, our model successfully identifies it in just 16 timesteps. Additionally, we observe a notable improvement in confidence scores for recognized objects; the confidence level for identifying a person, for instance, has surged from 0.47 to 0.88. In the area of semantic segmentation, AdaFire demonstrates an exceptional ability to delineate object boundaries accurately. A case in point is the enhanced clarity in capturing the giraffe's legs compared to the Calibration method. These results further demonstrate the effectiveness and generalizability of our proposed method.

\begin{figure}[t]
    \centering
    \begin{subfigure}[b]{0.3\textwidth}
        \centering
        \includegraphics[width=\textwidth]{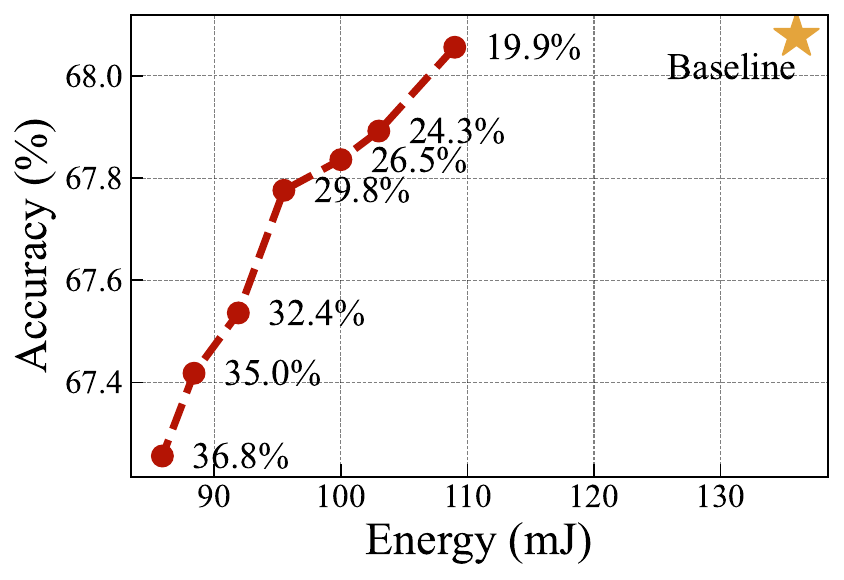} 
        \caption{ImageNet}
        \label{fig:imagenet}
    \end{subfigure}
    \begin{subfigure}[b]{0.3\textwidth}
        \centering
        \includegraphics[width=\textwidth]{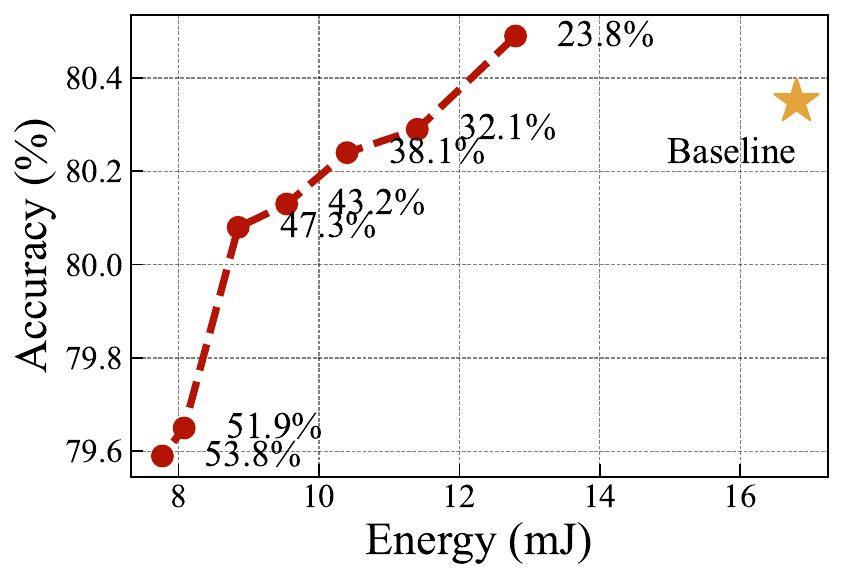} 
        \caption{CIFAR-100}
        \label{fig:cifar100}
    \end{subfigure}
    \begin{subfigure}[b]{0.3\textwidth}
        \centering
        \includegraphics[width=\textwidth]{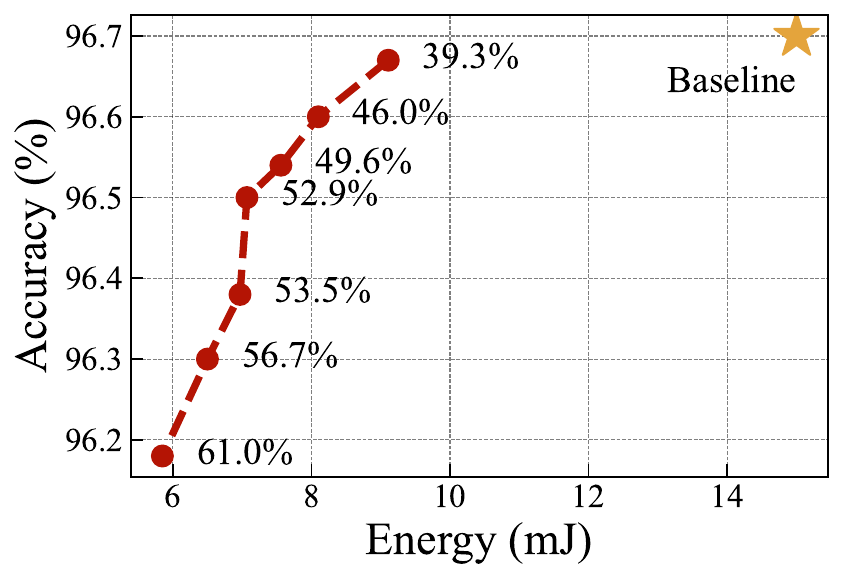} 
        \caption{CIFAR-10}
        \label{fig:cifar10}
    \end{subfigure}
    \vskip -0.1in
    \caption{\textbf{Effiveness of Sensitivity Spike Compression Technique (SSC)}. The baseline is the results without using SSC technique.}
    \vskip -0.1in
    \label{fig:SSC}
\end{figure}

\subsection{Effectiveness of Sensitivity Spike Compression Technique}
We evaluate the Sensitivity Spike Compression (SSC) technique on the CIFAR-10, CIFAR-100, and ImageNet datasets. As depicted in Fig.~\ref{fig:SSC}, we modulate \( S_{target} \) in accordance with Eq.~\ref{eq14}, observing its impact on both energy consumption and model performance. By default, \( S_{target} \) is set to reflect the cumulative sensitivity of the model prior to applying SSC. An increased value of \( S_{target} \) allocates a broader margin for energy reduction, though this may come at the cost of diminished accuracy. The empirical results highlight our method's capability to substantially diminish energy consumption with minimal trade-offs in performance. On the CIFAR-10 dataset, the technique achieves a remarkable \textbf{61.0\%} reduction in energy consumption, with only a minor decrease of 0.5\% in accuracy. Similarly, for the more compex ImageNet dataset, SSC results in a \textbf{32.4\%} energy saving, accompanied by a negligible 0.5\% drop in accuracy. These outcomes underscore the SSC technique's proficiency in enhancing energy efficiency without substantially compromising performance.

\subsection{Effectiveness of Input-aware Adaptive Timesteps Technique}

\begin{figure}[!t]
    \centering
    \begin{minipage}[t]{0.32\textwidth}
        \centering
        \includegraphics[width=\linewidth]{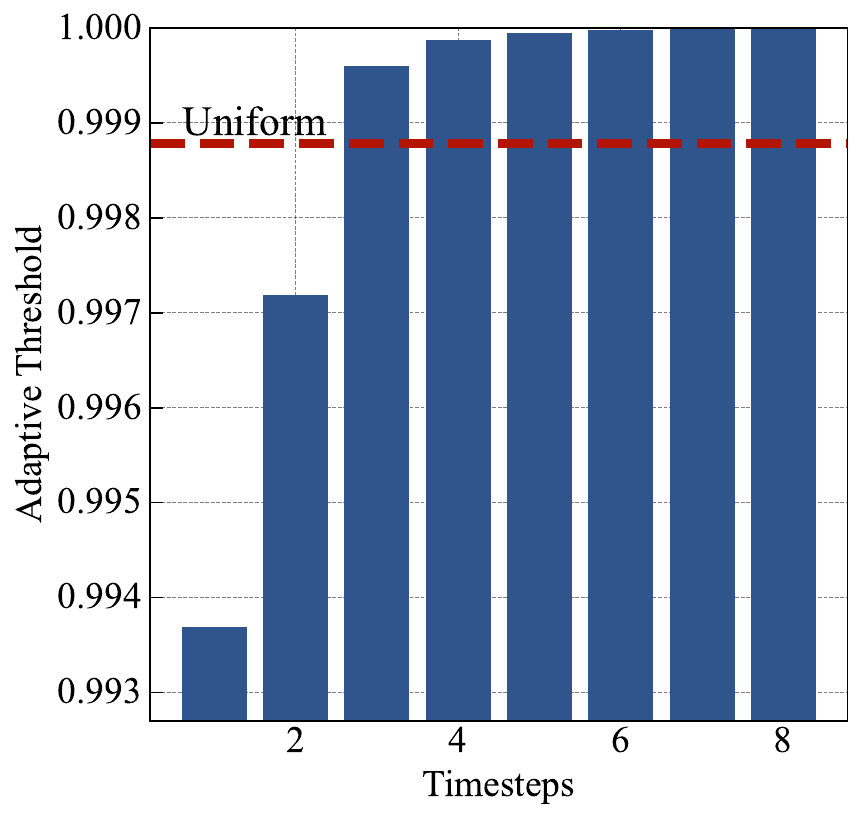}
        \subcaption{Adaptive v.s. Uniform.}
        \label{fig:uniform}
    \end{minipage}\hfill 
    \begin{minipage}[t]{0.66\textwidth}
        \centering
        \includegraphics[width=\linewidth]{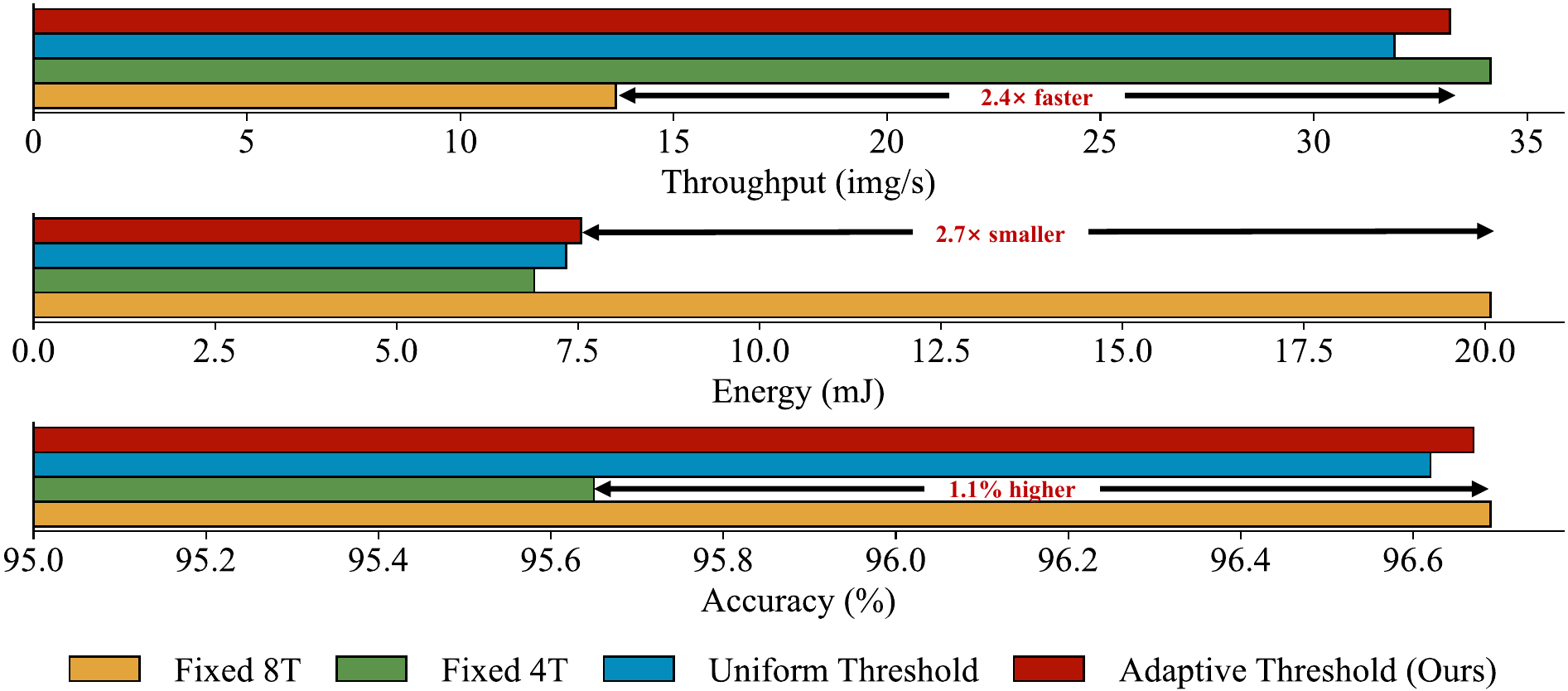}
        \subcaption{Performance comparison of different methods.}
        \label{fig:iat}
    \end{minipage}
    \vskip -0.05in
    \caption{Effectiveness of Input-aware Adaptive Timesteps Technique.}
    \vskip -0.05in
    \label{fig:IAT}
\end{figure}

We evaluate the effectiveness of the Input-aware Adaptive Timesteps (IAT) technique using the CIFAR-10 dataset. Fig.~\ref{fig:uniform} illustrates how our method dynamically adjusts the confidence score threshold to decide the optimal exit timestep. Unlike a uniform threshold approach, the IAT method progressively increases the threshold with each timestep. This adaptive strategy allows for early exit for simpler images, reducing latency, while allocating more processing time to complex images to preserve accuracy. As demonstrated in Fig.~\ref{fig:iat}, our IAT technique achieves a \textbf{2.4-fold} increase in speed and a \textbf{2.7-fold} reduction in energy consumption compared to the baseline method. Furthermore, when compared with the results using 3 timesteps, our method shows a performance improvement of \textbf{1.1\%}. These results underscore the IAT technique's capacity to significantly lower both latency and energy without compromising on performance.

\vskip -0.05in
\subsection{Ablation Study}

In this ablation study, we evaluate three proposed techniques—AdaFire, SSC, and IAT. We apply them to the CIFAR-10, CIFAR-100, and ImageNet datasets. Our results reveal that the AdaFire Neuron Model significantly boosts the accuracy of SNNs. Concurrently, the SSC and IAT techniques contribute to a substantial reduction in energy consumption. Remarkably, the synergistic application of three techniques leads to a groundbreaking \textbf{70.12\%} energy reduction and a 0.13\% accuracy enhancement for the CIFAR-10 dataset. For the more challenging ImageNet dataset, the combined implementation achieves a \textbf{43.10\%} decrease in energy usage while simultaneously enhancing accuracy by \textbf{11.53\%}. These results underscore the efficacy of our proposed conversion framework as a unified solution capable of both improving performance and efficiency.

\begin{table}[t]
\renewcommand{\arraystretch}{1}
    \setlength\tabcolsep{4pt} 
    \centering
\scriptsize
    \caption{Ablation Study of Different Techniques.}
    \vskip -0.1in
    \label{tab4}
\begin{tabular}{ccc|cc|cc|cc}
\hline
\multirow{2}{*}{\textbf{AdaFire}} & \multirow{2}{*}{\textbf{SSC}} & \multirow{2}{*}{\textbf{IAT}} & \multicolumn{2}{c}{\textbf{CIFAR-10}}         & \multicolumn{2}{c}{\textbf{CIFAR-100}}        & \multicolumn{2}{c}{\textbf{ImageNet}}         \\
                                  &                               &                               & \textbf{Acc. (\%)} & \textbf{Energy (mJ)} & \textbf{Acc. (\%)} & \textbf{Energy (mJ)} & \textbf{Acc. (\%)} & \textbf{Energy (mJ)} \\ \hline
                                  &                               &                               & 96.34                  & 14.86  (0)           & 79.90                   & 16.83 (0)            & 56.74                  & 162.56 (0)               \\
\cmark                                 &                               &                               & \textbf{96.69}                  & 20.49 (+37.88\%)    & \textbf{80.64}                  & 21.44 (+27.37\%)     & \textbf{68.45}                  & 169.52 (+0.04\%)     \\
\cmark                                 & \cmark                             &                               & 96.5                   & 7.71 (-48.12\%)      & 80.37                  & 10.00 (-40.58\%)        & 68.32                  & 120.32 (-25.98\%)       \\
\cmark                                 &                               & \cmark                             & 96.67                  & 7.06 (-52.48\%)      & 80.55                  & 12.16 (-27.75\%))    &       68.39                 &  85.15 (-47.61\%)                   \\
\cmark                                 & \cmark                             & \cmark                             & 95.47                  & \textbf{4.44 (-70.12\%)}      & 80.00                     & \textbf{6.69 (-60.25\%))}     &  68.27                      &     \textbf{92.50 (-43.10\%)}                
    \\\hline
    \end{tabular}
    \vskip -0.25in
\end{table}

\section{CONCLUSION}
In our paper, we propose a unified ANN-to-SNN conversion framework optimized for both performance and efficiency. We introduce the Adaptive-Firing Neuron Model (AdaFire), which significantly improves the SNN performance at low timesteps. Moreover, to improve efficiency, we propose a Sensitivity Spike Compression (SSC) technique that locates the adaptive threshold, and a Input-aware Adaptive Timesteps technique (ITA) that ajusts the timestep according to input, reducing both the energy consumption and latency of the conversion process. Collectively, these innovations present a unified conversion framework for enhancing the effectiveness and efficiency of SNNs.

\clearpage  

%
%
\bibliographystyle{splncs04}
\bibliography{main}

\begin{thebibliography}{10}
\providecommand{\url}[1]{\texttt{#1}}
\providecommand{\urlprefix}{URL }
\providecommand{\doi}[1]{https://doi.org/#1}

\bibitem{akopyanTruenorthDesignTool2015}
Akopyan, F., Sawada, J., Cassidy, A., {Alvarez-Icaza}, R., Arthur, J., Merolla, P., Imam, N., Nakamura, Y., Datta, P., Nam, G.J.: Truenorth: {{Design}} and tool flow of a 65 mw 1 million neuron programmable neurosynaptic chip. IEEE transactions on computer-aided design of integrated circuits and systems  \textbf{34}(10),  1537--1557 (2015)

\bibitem{buOptimalANNSNNConversion2021}
Bu, T., Fang, W., Ding, J., Dai, P., Yu, Z., Huang, T.: Optimal {{ANN-SNN Conversion}} for {{High-accuracy}} and {{Ultra-low-latency Spiking Neural Networks}}. In: International {{Conference}} on {{Learning Representations}} (2021)

\bibitem{buOptimalANNSNNConversion2021a}
Bu, T., Fang, W., Ding, J., Dai, P., Yu, Z., Huang, T.: Optimal {{ANN-SNN Conversion}} for {{High-accuracy}} and {{Ultra-low-latency Spiking Neural Networks}}. In: International {{Conference}} on {{Learning Representations}} (2021)

\bibitem{caiZeroqNovelZero2020}
Cai, Y., Yao, Z., Dong, Z., Gholami, A., Mahoney, M.W., Keutzer, K.: Zeroq: {{A}} novel zero shot quantization framework. In: Proceedings of the {{IEEE}}/{{CVF Conference}} on {{Computer Vision}} and {{Pattern Recognition}}. pp. 13169--13178 (2020)

\bibitem{cao2024spiking}
Cao, J., Wang, Z., Guo, H., Cheng, H., Zhang, Q., Xu, R.: Spiking denoising diffusion probabilistic models. In: Proceedings of the IEEE/CVF Winter Conference on Applications of Computer Vision. pp. 4912--4921 (2024)

\bibitem{cao2015spiking}
Cao, Y., Chen, Y., Khosla, D.: Spiking deep convolutional neural networks for energy-efficient object recognition. International Journal of Computer Vision  \textbf{113},  54--66 (2015)

\bibitem{connorsIntrinsicFiringPatterns1990}
Connors, B.W., Gutnick, M.J.: Intrinsic firing patterns of diverse neocortical neurons. Trends in neurosciences  \textbf{13}(3),  99--104 (1990)

\bibitem{danesh2019synaptic}
Danesh, C.D., Shaffer, C.M., Nathan, D., Shenoy, R., Tudor, A., Tadayon, M., Lin, Y., Chen, Y.: Synaptic resistors for concurrent inference and learning with high energy efficiency. Advanced Materials  \textbf{31}(18),  1808032 (2019)

\bibitem{daviesLoihiNeuromorphicManycore2018}
Davies, M., Srinivasa, N., Lin, T.H., Chinya, G., Cao, Y., Choday, S.H., Dimou, G., Joshi, P., Imam, N., Jain, S.: Loihi: {{A}} neuromorphic manycore processor with on-chip learning. Ieee Micro  \textbf{38}(1),  82--99 (2018)

\bibitem{davies2021advancing}
Davies, M., Wild, A., Orchard, G., Sandamirskaya, Y., Guerra, G.A.F., Joshi, P., Plank, P., Risbud, S.R.: Advancing neuromorphic computing with loihi: A survey of results and outlook. Proceedings of the IEEE  \textbf{109}(5),  911--934 (2021)

\bibitem{dengImagenetLargescaleHierarchical2009}
Deng, J., Dong, W., Socher, R., Li, L.J., Li, K., {Fei-Fei}, L.: Imagenet: {{A}} large-scale hierarchical image database. In: 2009 {{IEEE}} Conference on Computer Vision and Pattern Recognition. pp. 248--255. {Ieee} (2009)

\bibitem{dengOptimalConversionConventional2021}
Deng, S., Gu, S.: Optimal conversion of conventional artificial neural networks to spiking neural networks. arXiv preprint arXiv:2103.00476  (2021)

\bibitem{dengTemporalEfficientTraining2022}
Deng, S., Li, Y., Zhang, S., Gu, S.: Temporal {{Efficient Training}} of {{Spiking Neural Network}} via {{Gradient Re-weighting}}. arXiv preprint arXiv:2202.11946  (2022)

\bibitem{dingOptimalAnnsnnConversion2021}
Ding, J., Yu, Z., Tian, Y., Huang, T.: Optimal ann-snn conversion for fast and accurate inference in deep spiking neural networks. arXiv preprint arXiv:2105.11654  (2021)

\bibitem{everingham2010pascal}
Everingham, M., Van~Gool, L., Williams, C.K., Winn, J., Zisserman, A.: The pascal visual object classes (voc) challenge. International journal of computer vision  \textbf{88},  303--338 (2010)

\bibitem{fangIncorporatingLearnableMembrane2021}
Fang, W., Yu, Z., Chen, Y., Masquelier, T., Huang, T., Tian, Y.: Incorporating learnable membrane time constant to enhance learning of spiking neural networks. In: Proceedings of the {{IEEE}}/{{CVF International Conference}} on {{Computer Vision}}. pp. 2661--2671 (2021)

\bibitem{glatz2019adaptive}
Glatz, S., Martel, J., Kreiser, R., Qiao, N., Sandamirskaya, Y.: Adaptive motor control and learning in a spiking neural network realised on a mixed-signal neuromorphic processor. In: 2019 International Conference on Robotics and Automation (ICRA). pp. 9631--9637. IEEE (2019)

\bibitem{guSTCASpatiotemporalCredit2019}
Gu, P., Xiao, R., Pan, G., Tang, H.: {{STCA}}: {{Spatio-Temporal Credit Assignment}} with {{Delayed Feedback}} in {{Deep Spiking Neural Networks}}. In: Proceedings of the {{Twenty-Eighth International Joint Conference}} on {{Artificial Intelligence}}. pp. 1366--1372. {International Joint Conferences on Artificial Intelligence Organization}, {Macao, China} (Aug 2019). \doi{10.24963/ijcai.2019/189}

\bibitem{guo2017calibration}
Guo, C., Pleiss, G., Sun, Y., Weinberger, K.Q.: On calibration of modern neural networks. In: International conference on machine learning. pp. 1321--1330. PMLR (2017)

\bibitem{hoTCLANNtoSNNConversion2021}
Ho, N.D., Chang, I.J.: {{TCL}}: An {{ANN-to-SNN}} conversion with trainable clipping layers. In: 2021 58th {{ACM}}/{{IEEE Design Automation Conference}} ({{DAC}}). pp. 793--798. {IEEE} (2021)

\bibitem{izhikevichBurstsUnitNeural2003}
Izhikevich, E.M., Desai, N.S., Walcott, E.C., Hoppensteadt, F.C.: Bursts as a unit of neural information: Selective communication via resonance. Trends in neurosciences  \textbf{26}(3),  161--167 (2003)

\bibitem{kimFastAccurateObject2020}
Kim, S., Park, S., Na, B., Kim, J., Yoon, S.: Towards fast and accurate object detection in bio-inspired spiking neural networks through {{Bayesian}} optimization. IEEE Access  \textbf{9},  2633--2643 (2020)

\bibitem{kimSpikingyoloSpikingNeural2020}
Kim, S., Park, S., Na, B., Yoon, S.: Spiking-yolo: Spiking neural network for energy-efficient object detection. In: Proceedings of the {{AAAI}} Conference on Artificial Intelligence. vol.~34, pp. 11270--11277 (2020)

\bibitem{kimOptimizingDeeperSpiking2021a}
Kim, Y., Panda, P.: Optimizing deeper spiking neural networks for dynamic vision sensing. Neural Networks  \textbf{144},  686--698 (2021)

\bibitem{krahe2004burst}
Krahe, R., Gabbiani, F.: Burst firing in sensory systems. Nature Reviews Neuroscience  \textbf{5}(1),  13--23 (2004)

\bibitem{lan2023efficient}
Lan, Y., Zhang, Y., Ma, X., Qu, Y., Fu, Y.: Efficient converted spiking neural network for 3d and 2d classification. In: Proceedings of the IEEE/CVF International Conference on Computer Vision. pp. 9211--9220 (2023)

\bibitem{liEfficientAccurateConversion2022}
Li, Y., Zeng, Y.: Efficient and accurate conversion of spiking neural network with burst spikes. arXiv preprint arXiv:2204.13271  (2022)

\bibitem{liFreeLunchANN2021}
Li, Y., Deng, S., Dong, X., Gong, R., Gu, S.: A free lunch from {{ANN}}: {{Towards}} efficient, accurate spiking neural networks calibration. In: International {{Conference}} on {{Machine Learning}}. pp. 6316--6325. {PMLR} (2021)

\bibitem{liFreeLunchANN2021a}
Li, Y., Deng, S., Dong, X., Gong, R., Gu, S.: A free lunch from {{ANN}}: {{Towards}} efficient, accurate spiking neural networks calibration. In: International {{Conference}} on {{Machine Learning}}. pp. 6316--6325. {PMLR} (2021)

\bibitem{li2024seenn}
Li, Y., Geller, T., Kim, Y., Panda, P.: Seenn: Towards temporal spiking early exit neural networks. Advances in Neural Information Processing Systems  \textbf{36} (2024)

\bibitem{liDifferentiableSpikeRethinking2021}
Li, Y., Guo, Y., Zhang, S., Deng, S., Hai, Y., Gu, S.: Differentiable spike: {{Rethinking}} gradient-descent for training spiking neural networks. Advances in Neural Information Processing Systems  \textbf{34},  23426--23439 (2021)

\bibitem{liNeuromorphicDataAugmentation2022}
Li, Y., Kim, Y., Park, H., Geller, T., Panda, P.: Neuromorphic {{Data Augmentation}} for {{Training Spiking Neural Networks}}. arXiv preprint arXiv:2203.06145  (2022)

\bibitem{linMicrosoftCocoCommon2014}
Lin, T.Y., Maire, M., Belongie, S., Hays, J., Perona, P., Ramanan, D., Doll{\'a}r, P., Zitnick, C.L.: Microsoft coco: {{Common}} objects in context. In: Computer {{Vision}}\textendash{{ECCV}} 2014: 13th {{European Conference}}, {{Zurich}}, {{Switzerland}}, {{September}} 6-12, 2014, {{Proceedings}}, {{Part V}} 13. pp. 740--755. {Springer} (2014)

\bibitem{lismanBurstsUnitNeural1997}
Lisman, J.E.: Bursts as a unit of neural information: Making unreliable synapses reliable. Trends in neurosciences  \textbf{20}(1),  38--43 (1997)

\bibitem{liuEventbasedActionRecognition2021a}
Liu, Q., Xing, D., Tang, H., Ma, D., Pan, G.: Event-based {{Action Recognition Using Motion Information}} and {{Spiking Neural Networks}}. In: {{IJCAI}}. pp. 1743--1749 (2021)

\bibitem{maass1997networks}
Maass, W.: Networks of spiking neurons: the third generation of neural network models. Neural networks  \textbf{10}(9),  1659--1671 (1997)

\bibitem{mengTrainingHighPerformanceLowLatency2022}
Meng, Q., Xiao, M., Yan, S., Wang, Y., Lin, Z., Luo, Z.Q.: Training {{High-Performance Low-Latency Spiking Neural Networks}} by {{Differentiation}} on {{Spike Representation}}. In: Proceedings of the {{IEEE}}/{{CVF Conference}} on {{Computer Vision}} and {{Pattern Recognition}}. pp. 12444--12453 (2022)

\bibitem{mochizukiSimilarityNeuronalFiring2016}
Mochizuki, Y., Onaga, T., Shimazaki, H., Shimokawa, T., Tsubo, Y., Kimura, R., Saiki, A., Sakai, Y., Isomura, Y., Fujisawa, S.: Similarity in neuronal firing regimes across mammalian species. Journal of Neuroscience  \textbf{36}(21),  5736--5747 (2016)

\bibitem{neftciSurrogateGradientLearning2019}
Neftci, E.O., Mostafa, H., Zenke, F.: Surrogate {{Gradient Learning}} in {{Spiking Neural Networks}}: {{Bringing}} the {{Power}} of {{Gradient-Based Optimization}} to {{Spiking Neural Networks}}. IEEE Signal Processing Magazine  \textbf{36}(6),  51--63 (Nov 2019). \doi{10.1109/MSP.2019.2931595}

\bibitem{park2019fast}
Park, S., Kim, S., Choe, H., Yoon, S.: Fast and efficient information transmission with burst spikes in deep spiking neural networks. In: Proceedings of the 56th Annual Design Automation Conference 2019. pp.~1--6 (2019)

\bibitem{qi2017pointnet}
Qi, C.R., Su, H., Mo, K., Guibas, L.J.: Pointnet: Deep learning on point sets for 3d classification and segmentation. In: Proceedings of the IEEE conference on computer vision and pattern recognition. pp. 652--660 (2017)

\bibitem{roy2019towards}
Roy, K., Jaiswal, A., Panda, P.: Towards spike-based machine intelligence with neuromorphic computing. Nature  \textbf{575}(7784),  607--617 (2019)

\bibitem{stocklOptimizedSpikingNeurons2021}
St{\"o}ckl, C., Maass, W.: Optimized spiking neurons can classify images with high accuracy through temporal coding with two spikes. Nature Machine Intelligence  \textbf{3}(3),  230--238 (2021)

\bibitem{teerapittayanon2016branchynet}
Teerapittayanon, S., McDanel, B., Kung, H.T.: Branchynet: Fast inference via early exiting from deep neural networks. In: 2016 23rd international conference on pattern recognition (ICPR). pp. 2464--2469. IEEE (2016)

\bibitem{versace2010brain}
Versace, M., Chandler, B.: The brain of a new machine. IEEE spectrum  \textbf{47}(12),  30--37 (2010)

\bibitem{vialeCarsnnEfficientSpiking2021}
Viale, A., Marchisio, A., Martina, M., Masera, G., Shafique, M.: Carsnn: {{An}} efficient spiking neural network for event-based autonomous cars on the loihi neuromorphic research processor. In: 2021 {{International Joint Conference}} on {{Neural Networks}} ({{IJCNN}}). pp. 1--10. {IEEE} (2021)

\bibitem{vitale2021event}
Vitale, A., Renner, A., Nauer, C., Scaramuzza, D., Sandamirskaya, Y.: Event-driven vision and control for uavs on a neuromorphic chip. In: 2021 IEEE International Conference on Robotics and Automation (ICRA). pp. 103--109. IEEE (2021)

\bibitem{wangSignedNeuronMemory2022}
Wang, Y., Zhang, M., Chen, Y., Qu, H.: Signed neuron with memory: {{Towards}} simple, accurate and high-efficient ann-snn conversion. In: International {{Joint Conference}} on {{Artificial Intelligence}} (2022)

\bibitem{wang2023masked}
Wang, Z., Fang, Y., Cao, J., Zhang, Q., Wang, Z., Xu, R.: Masked spiking transformer. In: Proceedings of the IEEE/CVF International Conference on Computer Vision. pp. 1761--1771 (2023)

\bibitem{yaoTemporalwiseAttentionSpiking2021}
Yao, M., Gao, H., Zhao, G., Wang, D., Lin, Y., Yang, Z., Li, G.: Temporal-wise attention spiking neural networks for event streams classification. In: Proceedings of the {{IEEE}}/{{CVF International Conference}} on {{Computer Vision}}. pp. 10221--10230 (2021)

\bibitem{yi2016scalable}
Yi, L., Kim, V.G., Ceylan, D., Shen, I.C., Yan, M., Su, H., Lu, C., Huang, Q., Sheffer, A., Guibas, L.: A scalable active framework for region annotation in 3d shape collections. ACM Transactions on Graphics (ToG)  \textbf{35}(6),  1--12 (2016)

\bibitem{zeldenrustNeuralCodingBursts2018}
Zeldenrust, F., Wadman, W.J., Englitz, B.: Neural {{Coding With Bursts}}\textemdash{{Current State}} and {{Future Perspectives}}. Frontiers in Computational Neuroscience  \textbf{12} (2018)

\bibitem{zhangSpikingTransformersEventBased2022}
Zhang, J., Dong, B., Zhang, H., Ding, J., Heide, F., Yin, B., Yang, X.: Spiking {{Transformers}} for {{Event-Based Single Object Tracking}}. In: Proceedings of the {{IEEE}}/{{CVF Conference}} on {{Computer Vision}} and {{Pattern Recognition}}. pp. 8801--8810 (2022)

\bibitem{zhouSpikformerWhenSpiking2022}
Zhou, Z., Zhu, Y., He, C., Wang, Y., Yan, S., Tian, Y., Yuan, L.: Spikformer: {{When Spiking Neural Network Meets Transformer}}. arXiv preprint arXiv:2209.15425  (2022)

\end{thebibliography}
\end{document}